# Technical Note: Exploring $\Sigma_2^P$ / $\Pi_2^P$-hardness for Argumentation Problems with fixed distance to tractable classes.

Wolfgang Dvořák[*]

Institute of Information Systems 184/2, Technische Universität Wien,

dvorak@dbai.tuwien.ac.at

**Abstract**

We study the complexity of reasoning in abstracts argumentation frameworks close to a graph classes that allow for efficient reasoning methods, i.e. to one of the classes of acyclic, noeven, bipartite and symmetric AFs. In this work we show that certain reasoning problems on the second level of the polynomial hierarchy still maintain their full complexity when restricted to instances of fixed distance to one of the above graph classes.

## 1 Overview

This work complements the studies in [9] on augmenting tractable fragments of abstract argumentation, but in contrast solely addresses negative results.

That is we consider abstracts argumentation which are close to a graph classes which allows for efficient reasoning methods, i.e. to one of the classes of acyclic [4], noeven [6], bipartite [5] and symmetric [2] AFs. We show that certain reasoning problems on the second level of the polynomial hierarchy still maintain their full complexity when restricted to instances of fixed distance to one of the above graph classes. This improves results from [9], showing hardness for the

[*]This work has been funded by the Vienna Science and Technology Fund (WWTF) through project ICT08-028.

|              | acyclic | noeven | bipartite | symmetric |
|--------------|---------|--------|-----------|-----------|
| $Skept_{prf}$ | fpt     | fpt    | 1         | $\leq 2$  |
| $Cred_{sem}$  | fpt     | fpt    | 1         | $\leq 2$  |
| $Skept_{sem}$ | fpt     | fpt    | 1         | $\leq 2$  |
| $Cred_{stg}$  | 1       | 0      | $\leq 4$  | $\leq 4$  |
| $Skept_{stg}$ | 1       | 0      | $\leq 4$  | $\leq 4$  |

Table 1: Distances for second level hardness

.

first level of the polynomial hierarchy and therefore that certain tractable graph classes do not maintain an augmentation w.r.t. the distance to a graph class.

An overview of our results, together with fixed-parameter tractability results from [9], is given in Table 1. An entry $\leq k$ encodes that the respective reasoning problem is hard for the second level of the polynomial hierarchy, i.e. either $\Pi_2^P$ or $\Sigma_2^P$ hard, even when restricted to instances with distance to the specific graph class $\leq k$.[1] Respective for an entry 1 the problems maintain there full hardness even for instances with distance 1. An entry "fpt" denotes that the problem is fixed-parameter tractable (cf. [9]) w.r.t. the distance to the specific graph class, which implies that there is no constant maintaining (full) hardness. Moreover for stage semantics, the class of noeven itself bears the full complexity, which is mirrored by the distance 0 entries in the table. Finally let us mention that [9] provides NP / coNP hardness results even for distance 1 to symmetric.

The remaining of the paper is organised as follows:

- Section 2 introduces the necessary background. In particular we give the definition of argumentation frameworks and the related semantics.

- In Section 3 we present our technical results, i.e. reductions showing hardness for specific graph classes.

---

[1] We do not claim that these distances are optimal, it might be the case that ever lower distances suffices to obtain the full hardness.

## 2 Preliminaries

In this section we introduce (abstract) argumentation frameworks [4] and recall the definitions of the semantics we study in this paper.

**Definition 1.** *An* argumentation framework (AF) *is a pair $F = (A, R)$ where $A$ is a set of arguments and $R \subseteq A \times A$ is the attack relation. For a given AF $F = (A, R)$ we use $A_F$ to denote the set $A$ of its arguments and $R_F$ to denote its attack relation $R$. The pair $(a, b) \in R$ means that $a$ attacks $b$. We sometimes use the notation $a \rightarrowtail^R b$ instead of $(a, b) \in R$. For $S \subseteq A$ and $a \in A$, we also write $S \rightarrowtail^R a$ (resp. $a \rightarrowtail^R S$) in case there exists an argument $b \in S$, such that $b \rightarrowtail^R a$ (resp. $a \rightarrowtail^R b$). In case no ambiguity arises, we use $\rightarrowtail$ instead of $\rightarrowtail^R$.*

Semantics for argumentation frameworks are given via a function $\sigma$ which assigns to each AF $F = (A, R)$ a set $\sigma(F) \subseteq 2^A$ of extensions. We shall consider here for $\sigma$ the functions $stb$, $adm$, $prf$, $com$, $grd$, $stg$, and $sem$ which stand for stable, admissible, preferred, complete, grounded, stage, and respectively, semi-stable semantics. Before giving the actual definitions for these semantics, we require a few more formal concepts.

**Definition 2.** *Given an AF $F = (A, R)$, an argument $a \in A$ is* defended *(in $F$), by a set $S \subseteq A$ if for each $b \in A$, such that $b \rightarrowtail a$, also $S \rightarrowtail b$ holds. Moreover, for a set $S \subseteq A$, we define the* range of $S$, *denoted as $S_R^+$, as the set $S \cup \{b \mid S \rightarrowtail b\}$. We write $S \leq_R^+ E$ iff $S_R^+ \subseteq E_R^+$.*

We continue with the definitions of argumentation semantics.

**Definition 3.** *Let $F = (A, R)$ be an AF. A set $S \subseteq A$ is* conflict-free *(in $F$), if there are no $a, b \in S$, such that $(a, b) \in R$. For such a conflict-free set $S$, it holds that*

- $S \in stb(F)$, *if $S_R^+ = A$;*

- $S \in adm(F)$, *if each $a \in S$ is defended by $S$;*

- $S \in prf(F)$, *if $S \in adm(F)$ and there is no $T \in adm(F)$ with $T \supset S$;*

- $S \in com(F)$, *if $S \in adm(F)$ and for each $a \in A$ that is defended by $S$, $a \in S$;*

- $S \in grd(F)$, *if $S \in com(F)$ and there is no $T \in com(F)$ with $T \subset S$;*

| $\sigma$ | $Cred_\sigma$ | $Skept_\sigma$ |
|---|---|---|
| $prf$ | NP-complete | $\Pi_2^P$-complete |
| $sem$ | $\Sigma_2^P$-complete | $\Pi_2^P$-complete |
| $stg$ | $\Sigma_2^P$-complete | $\Pi_2^P$-complete |

Table 2: Complexity of credulous and skeptical acceptance for the semantics under our considerations.

- $S \in stg(F)$, if there is no conflict-free set $T$ in $F$, such that $T_R^+ \supset S_R^+$;

- $S \in sem(F)$, if $S \in adm(F)$ and there is no $T \in adm(F)$ with $T_R^+ \supset S_R^+$.

We recall that for each AF $F$, $stb(F) \subseteq sem(F) \subseteq prf(F) \subseteq com(F) \subseteq adm(F)$ holds, and that for each of the considered semantics $\sigma$ except stable semantics, $\sigma(F) \neq \emptyset$ holds. The grounded semantics always yields exactly one extension. Moreover if an AF has at least one stable extension then its stable, semi-stable, and stage extensions coincide.

Next we briefly recall results concerning the complexity of reasoning. We assume the reader is familiar with standard complexity theory and in particular with the polynomial hierarchy (see e.g. [10]). We are interested in the the following decision problems for the semantics $\sigma$ on the second level of the polynomial hierarchy:

- *Credulous Acceptance* $Cred_\sigma$: Given AF $F = (A, R)$ and an argument $a \in A$. Is $a$ contained in some $S \in \sigma(F)$?

- *Skeptical Acceptance* $Skept_\sigma$: Given AF $F = (A, R)$ and an argument $a \in A$. Is $a$ contained in each $S \in \sigma(F)$?

We summarize the general complexity of the mentioned reasoning problems [1, 3, 6, 7] in Table 2.

Finally we introduce the *distance to graph class* which is closely related to the notation of a backdoor (see [9]).

**Definition 4.** *Let $\mathcal{G}$ be a graph class and $F = (A, R)$ an AF. We define $\mathrm{dist}_\mathcal{G}(F)$ as the minimal number $k$ such that there exists a set $S \subseteq A$ with $|S| = k$ and $(A \setminus S, R \cap (A \setminus S \times A \setminus S)) \in \mathcal{G}$. If there is no such set $S$ we define $\mathrm{dist}_\mathcal{G}(F) = \infty$.*

Following [9], we study the graph classes of acyclic (ACY), even cycle-free (NOEVEN), symmetric (SYM) and bipartite (BIP) graphs. In particular we consider decision problems which are not fixed-parameter tractable w.r.t. the above introduced distance to a fragment and prove that the problems are hard for the second level of the polynomial hierarchy even for a fixed distance to a tractable fragment.

## 3  Technical Results

Most of the results in this section build on reductions from deciding whether a quantified Boolean formula (QBF) in a particular form is valid. More concrete we consider $QBF_2^\forall$ formulae, which are of the form $\forall Y \exists X\, \varphi$ where $X$ and $Y$ are strings of propositional atoms and $\varphi$ is a propositional formula over the atoms $X \cup Y$ (we may assume that $\varphi$ is in 3-CNF, $\varphi$ is satisfiable or $\varphi$ is monotone). We say that a QBF $\Phi = \forall Y \exists X\, \varphi$ is *valid* if for each $M_Y \subseteq Y$ there exists an $M_X \subseteq X$ such that $M = M_Y \cup M_X$ is a model of $\varphi$. The problem $QSAT_2^\forall$, deciding whether a given $QBF_2^\forall$ formula is valid, is well-known to be $\Pi_2^P$-complete.

**Reduction 1.** *Given a QBF $\Phi = \forall Y \exists X\, \varphi$ with $\varphi$ being a monotone CNF, $C$ being the set of positive clauses, $\bar{C}$ being the set of negative clauses and $X = Y \cup Z$. The AF $F_\Phi^1 = (A, R)$ is constructed as follows:*

$$
\begin{aligned}
A \;=\;& \{\varphi, b, \bar{b}\} \cup C \cup \bar{C} \cup X \cup \bar{X} \\
R \;=\;& \{(c, \varphi) \mid c \in C \cup \bar{C}\} \cup \{(\varphi, b), (\varphi, \bar{b})\} \cup \\
& \{(x, \bar{x}), (\bar{x}, x) \mid x \in X\} \cup \\
& \{(l, c), (c, l) \mid \text{literal } l \text{ occurs in } c \in C \cup \bar{C}\} \cup \\
& \{(b, c) \mid c \in C\} \cup \{(\bar{b}, \bar{c}) \mid \bar{c} \in \bar{C}\} \cup \\
& \{(b, \bar{x}) \mid \bar{x} \in \bar{X}\} \cup \{(\bar{b}, x) \mid x \in X\}
\end{aligned}
$$

For an illustration of this reduction see Figure 1.

**Lemma 1.** *For a monotone $QSAT_\forall^2$ formula $\Phi$ and the AF $F_\Phi^1$ the following holds:*

1. *The arguments $\bar{b}, b, c \in C \cup \overline{C}$ are not contained in any admissible set of $F_\Phi^1$.*

2. *For each set $S \subseteq Y$ the set $S \cup \overline{Y \setminus S}$ is admissible in $F_\Phi^1$ and each set $E \subsetneq Y \cup \overline{Y}$ not of this form is not a preferred extension.*

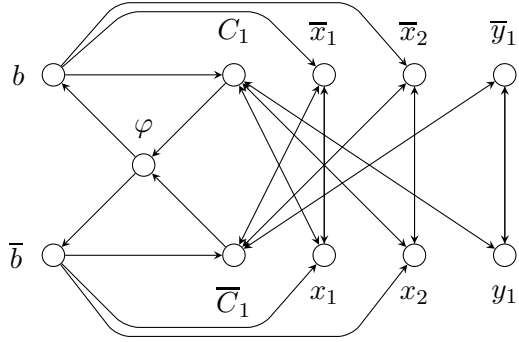

Figure 1: Illustration of the AF $F_\Phi^2$ obtained from the monotone $QSAT_\forall^2$ formula $\Phi = \forall y_1 \exists x_1 \exists x_2 C_1 \wedge \overline{C}_1$ with $C_1 = x_1 \vee x_2 \vee y_1$ and $\overline{C}_1 = \neg x_1 \vee \neg x_2 \vee \neg y_1$.

3. For any preferred extension $E$ of $F_\Phi^1$, if $\Phi \notin E$ then $E = S \cup \overline{Y \setminus S}$ for some $S \subseteq Y$.

4. For any preferred extension $E$ of $F_\Phi^1$, if $\Phi \in E$ then $E \cap (X \cup Y)$ is a model of $\varphi$.

5. If $M$ is a model of $\varphi$ then $M \cup \overline{(X \cup Y) \setminus M} \cup \{\Phi\}$ is a preferred extension of $F_\Phi^1$.

*Proof.* We prove each point separately:

1. In order to obtain a contradiction let us assume that there exists an admissible set $E$ containing an argument $c \in C \cup \overline{C}$. We have that $c$ is either attacked by $\overline{b}$ or $b$, and as $E$ is an admissible set it must contain an argument that defends $c$. However, $\Phi$ is the only argument that attacks $\overline{b}$ or $b$, thus $\Phi \in E$. But as $c$ attacks $\Phi$ this contradicts the conflict-freeness of $E$. Hence we conclude that no admissible set of $F$ contains an argument from $C \cup \overline{C}$.

   Next let us assume that there exists an admissible set $E$ containing $\overline{b}$ or $b$. Then $E$ defends $\overline{b}$ or $b$, respectively, and thus contains an argument attacking $\Phi$. As the only arguments attacking $\Phi$ are those in the set $C \cup \overline{C}$ this contradicts the above observation.

2. As all attacks concerning arguments in $Y \cup \overline{Y}$ are mutual attacks we conclude that a subset of $Y \cup \overline{Y}$ is admissible if and only if it is conflict-free. One can see that the maximal conflict free subsets of $Y \cup \overline{Y}$ are the sets $S \cup \overline{Y \setminus S}$ with $S \subseteq Y$. Hence we can conclude that (i) these sets are admissible, (ii) each admissible subset of $Y \cup \overline{Y}$ is either of the form $S \cup \overline{Y \setminus S}$ with $S \subseteq Y$ or the subset of such a set, and (iii) no subset of such a set can be maximal admissible, i.e., preferred.

3. As mentioned in (1) the arguments $\overline{b}, b, c \in C \cup \overline{C}$ are not contained in any admissible set. Further, since $\Phi \notin E$ we obtain that the arguments $x \in X \cup \overline{X}$ are not defended by $E$ and thus not contained in $E$. The only arguments that are left belong to $Y \cup \overline{Y}$ and hence by (2) $E = S \cup \overline{Y \setminus S}$.

4. As $\Phi \in E$ it follows that each argument in $C \cup \overline{C}$ is attacked by $E$. Further as $\overline{b}, b \notin E$ each argument in $C \cup \overline{C}$ is attacked by an argument in $E \cap (X \cup Y \cup \overline{X} \cup \overline{Y})$. Thus by construction, $E \cap (X \cup Y)$ is a model of $\varphi$.

5. Clearly $M \cup \overline{(X \cup Y) \setminus M} \cup \{\Phi\}$ is conflict-free. We have mutual attacks between arguments $x \in X \cup Y$ and the corresponding arguments $\overline{x} \in \overline{X} \cup \overline{Y}$. Hence all arguments in $X \cup Y \cup \overline{X} \cup \overline{Y}$ are either in the set $M \cup \overline{(X \cup Y) \setminus M}$ or attacked by some argument from this set. Further, as $M$ is a model of $\Phi$, it follows by construction that the arguments $M \cup \overline{(X \cup Y) \setminus M}$ attack all the arguments in $C \cup \overline{C}$ and thus defend $\Phi$. Finally the argument $\Phi$ attacks both $b$ and $\overline{b}$. That is, each argument of $F$ either is in the set $M \cup \overline{(X \cup Y) \setminus M} \cup \{\Phi\}$ or attacked by an argument from this set. Hence the set $M \cup \overline{(X \cup Y) \setminus M} \cup \{\Phi\}$ is a stable extension of $F$, and thus also an preferred extension of $F$.

$\square$

**Proposition 1.** *A monotone $QSAT_\forall^2$ formula $\Phi$ is valid iff the argument $\Phi$ is skeptically accepted in $F_\Phi^1$ with respect to $prf$.*

*Proof.* We are first going to show that the formula $\Phi$ is valid only if the argument $\Phi$ is skeptically accepted in $F$ with respect to $prf$. To this end we consider a valid formula $\Phi = \forall Y \exists X \varphi(X, Y)$. In order to obtain a contradiction let us assume that there exists a preferred extension $E$ such that $\Phi \notin E$. Then we have that $E = S \cup \overline{Y \setminus S}$ for some $S \subseteq Y$. Using that the formula $\Phi$ is valid, we conclude that there exists a model $M$ of $\varphi$ such that $S \subsetneq M$. But then $E' = M \cup \overline{(X \cup Y) \setminus M} \cup \{\Phi\}$ is a preferred extension of $F$ and $E \subsetneq E'$, a contradiction.

It remains to show that the formula $\Phi$ is valid if the argument $\Phi$ is skeptically accepted in $F$ with respect to $prf$. To this end let us assume that $\Phi$ is not valid, i.e., there exists an $S \subsetneq Y$ which is not contained in any model of $\varphi$. Now let us consider an arbitrary preferred extension $E$ such that $S \cup \overline{Y \setminus S} \subseteq E$. Such an $E$ must exist as $S \cup \overline{Y \setminus S}$ is an admissible set. In order to obtain a contradiction we assume that $\Phi \in E$. It follows that $E \cap (X \cup Y)$ is a model of $\varphi$ containing $S$, a contradiction. Thus $\Phi \notin E$ and we conclude that $E$ is a preferred extension that does not contain $\Phi$. Hence the argument $\Phi$ is not skeptically accepted in $F$ with respect to the preferred semantics. □

**Theorem 1.** *The problem $Skept_{prf}$ is $\Pi_2^P$-complete even for AFs $F$ with $\text{dist}_{BIP}(F) = 1$.*

*Proof.* By Proposition 1 one can use $F_\Phi^1$ to reduce the $\Pi_2^P$-hard problem, deciding whether a monotone $QSAT_\forall^2$ formula is valid to $Skept_{prf}$. Now mention that deleting the argument $\varphi$ from the AF $F_\Phi^1$ would result a bipartite AF. □

**Reduction 2.** *Given a QBF $\Phi = \forall Y \exists X \, \varphi$ with $\varphi$ being a CNF, $C$ being the set of clauses and $X = Y \cup Z$. The AF $F_\Phi^2 = (A, R)$ is constructed as follows:*

$$\begin{aligned}
A &= \{\varphi, b\} \cup C \cup \bar{C} \cup X \cup \bar{X} \\
R &= \{(c, \varphi) \mid c \in C\} \cup \{(\varphi, b)\} \cup \\
&\quad \{(x, \bar{x}), (\bar{x}, x) \mid x \in X\} \cup \\
&\quad \{(l, c), (c, l) \mid \text{literal } l \text{ occurs in } c \in C\} \cup \\
&\quad \{(b, c) \mid c \in C \cup \bar{C}\} \cup \{(b, x) \mid x \in X \cup \bar{X}\}
\end{aligned}$$

**Theorem 2.** *The problem $Skept_{prf}$ is $\Pi_2^P$-complete even for AFs $F$ with $\text{dist}_{SYM}(F) = 2$.*

*Proof sketch.* One can show that $F_\Phi^2$ (similar to Proposition 1) reduces the $\Pi_2^P$-hard problem, deciding whether a $QSAT_\forall^2$ formula is valid to $Skept_{prf}$. Furthermore deleting the arguments $\varphi, b$ from the AF $F_\Phi^2$ would result a symmetric AF. □

**Reduction 3.** *Given a QBF $\Phi = \forall Y \exists Z \, \varphi$ with $\varphi$ being a monotone CNF, $C$ being the set of positive clauses, $\bar{C}$ being the set of negative clauses and $X = Y \cup Z$. The AF $F_\Phi^3 = (A, R)$ is constructed as follows:*

$$\begin{aligned}
A_\Phi &= \{\varphi_p, \varphi_n, \bar{\varphi}, b, g\} \cup C \cup \bar{C} \cup X \cup \bar{X} \cup Y' \cup \bar{Y}' \\
R_\Phi &= \{\langle c, \varphi_p\rangle \mid c \in C\} \cup \{\langle \bar{c}, \varphi_p\rangle \mid \bar{c} \in \bar{C}\} \cup \{\langle x, \bar{x}\rangle, \langle \bar{x}, x\rangle \mid x \in X\} \cup \\
&\quad \{\langle x, c\rangle, \langle c, x\rangle \mid x \text{ occurs in } c\} \cup \{\langle \bar{x}, c\rangle, \langle c, \bar{x}\rangle \mid \neg x \text{ occurs in } c\} \\
&\quad \{\langle y, y'\rangle, \langle y', y\rangle, \langle \bar{y}, \bar{y}'\rangle, \langle \bar{y}', \bar{y}\rangle \mid y \in Y\} \cup \{(g, y'), (g, \bar{y}') \mid y \in Y\} \cup \\
&\quad \{\langle \varphi_p, b\rangle, \langle b, \varphi_p\rangle, \langle g, g\rangle, \langle g, b\rangle\} \cup \{(g, c) \mid c \in C\} \cup \{(\varphi_p, \bar{\varphi}), (\bar{\varphi}, \varphi_p)\}
\end{aligned}$$

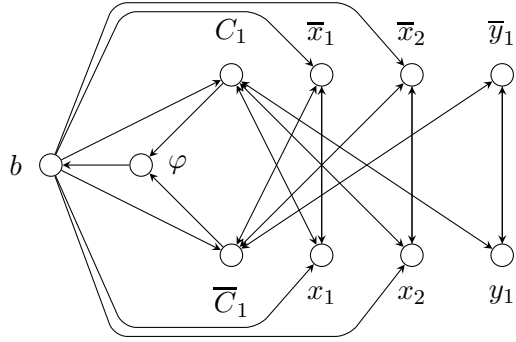

Figure 2: Illustration of the AF $F_\Phi^2$ obtained from the $QSAT_\forall^2$ formula $\Phi = \forall y_1 \exists x_1 \exists x_2 C_1 \wedge \overline{C}_1$ with $C_1 = x_1 \vee x_2 \vee y_1$ and $\overline{C}_1 = \neg x_1 \vee \neg x_2 \vee \neg y_1$.

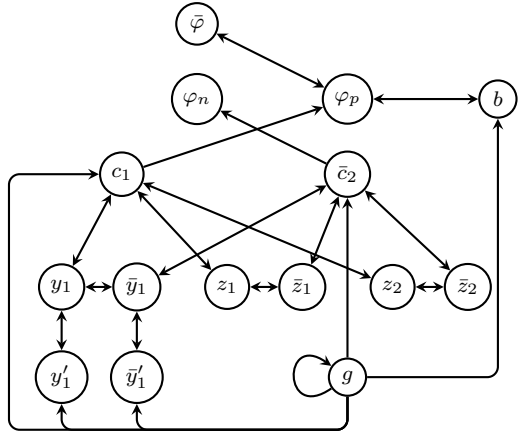

Figure 3: Illustration of the AF $F_\Phi^3$ obtained from the monotone $QSAT_\forall^2$ formula $\Phi = \forall y_1 \exists z_1 \exists z_2 C_1 \wedge \overline{C}_1$ with $C_1 = x_1 \vee x_2 \vee y_1$ and $\overline{C}_1 = \neg x_1 \vee \neg x_2 \vee \neg y_1$.

For an example, see Figure 3.

**Proposition 2.** *For a monotone QBF* $\Phi = \forall Y \exists Z \varphi$ *with each clause containing an literal from* $Z \cup \bar{Z}$, *the following statements are equivalent*

1. $\Phi$ *is valid.*

2. $\varphi_p$ *is skeptically accepted in* $F_\Phi^3$ *w.r.t. semi-stable semantics.*

3. $\bar\varphi$ *is not credulously accepted in* $F_\Phi^3$ *w.r.t. semi-stable semantics.*

*Proof sketch.* (1) $\Leftrightarrow$ (2): Notice that this reduction is a variation of the reduction presented in [7]. Recall that the candidates for being semi-stable extensions are the preferred extensions. We have that none of the arguments $\{g, b\} \cup C \cup \bar{C} \cup Y' \cup \bar{Y}'$ is acceptable for a semi-stable extension, as none of them can be defended. Moreover each $E, E_1, E_2 \in prf(F_\Phi^3)$ satisfies:

- $y \in Y \Rightarrow |\{y, \bar{y}\} \cap E| = 1$
- $z \in Z \Rightarrow |\{z, \bar{z}\} \cap E| = 1$
- For $y \in Y$ if $y \in E_1$ and $\bar{y} \in E_2$ then $E_1 \not\leq^+ E_2$ and $E_2 \not\leq^+ E_1$.

So each preferred extension corresponds to an true-assignment on $Y \cup Z$, and each true-assignment to at least on preferred extension. Further we have that assignments that differ on $Y$, result preferred extensions with incomparable range.

$\Rightarrow$: Let us assume that $\Phi = \forall Y \exists Z \varphi$ is valid and let us consider an $E \in sem(F_\Phi^3)$. $E$ give rise to an true-assignment $M_Y = E \cap Y$ on variables $Y$, by assumption we have that there exists $M_Z \subseteq Z$ such that $M_Y \cup M_Z$ is a model of $\varphi$. Then the set $G = M_Y \cup \overline{Y \setminus M_Y} \cup M_Z \cup \overline{Z \setminus M_Z} \cup \{\varphi_p, \varphi_n\}$ is an admissible set with $G^+ = A \setminus (\{g\} \cup (Y \setminus M_Y') \cup \bar{M}_Y{'})$. One can easily show that no admissible set containing $M_Y \cup \overline{Y \setminus M_Y}$ has any of the arguments $(\{g\} \cup (Y \setminus M_Y') \cup \bar{M}_Y{'})$ it its range and hence $G$ is $\leq^+$-maximal. As by assumption also $E$ is $\leq^+$-maximal we have $E^+ = G^+$ and as $b \in E^+$ we have $\varphi \in E$.

$\Leftarrow$: Let us assume that $\Phi = \forall Y \exists Z \varphi$ is not valid, i.e. there is an $M_Y \subseteq Y$ such that there is no $M_Z \subseteq Z$ such that $M_Y \cup M_Z$ is a model of $\varphi$. We consider the set $E = M_Y \cup \overline{Y \setminus M_Y} \cup \bar{Z} \cup \{\bar{\varphi}, \varphi_n\}$ which is admissible as each clause of $\bar{C}$ contains at least one literal from $\bar{Z}$. We have that $E^+ = A \setminus (\{g, b\} \cup (Y \setminus M_Y') \cup \bar{M}_Y{'})$. Hence we have that $E$ is semi-stable unless there exists an extension $G$ with $G^+ = A \setminus (\{g\} \cup (Y \setminus M_Y') \cup \bar{M}_Y{'})$. Such a $G$ must contain both $\varphi_n$ and $\varphi_p$ and thus attacks each $c \in C \cup \bar{C}$. Hence, by construction of $F_\Phi^3$, $G$ would corresponds to a model $M_Y \cup M_Z$ of $\varphi$, a contradiction to our assumption that there is no $M_Z \subseteq Z$. Thus $E$ is a semi-stable extension with $\varphi \notin E$.

(2) $\Leftrightarrow$ (3): As $\bar{\varphi}$ is only in conflict with $\varphi_p$ for each $E \in sem(F_\Phi^3)$ either $\varphi_p \in E$ or $\bar{\varphi} \in E$ holds. $\square$

**Theorem 3.** *The problem $Cred_{sem}$ is $\Sigma_2^P$-complete and the problem $Skept_{sem}$ is $\Pi_2^P$-complete even for AFs $F$ with $\text{dist}_{BIP}(F) \leq 1$.*

*Proof.* By Proposition 2 and the observation that for monotone QBFs the $F_\Phi^3 - \{g\}$ is bipartite. $\square$

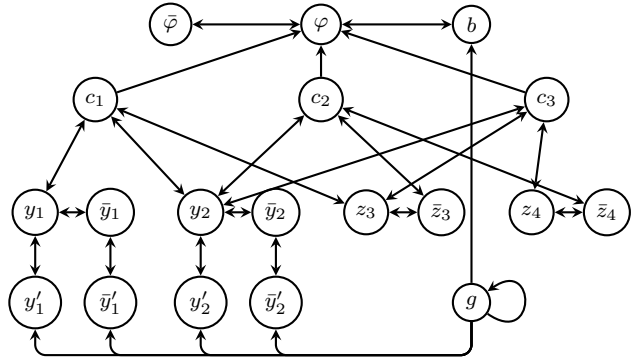

Figure 4: $F_\Phi^4$ for the QBF $\Phi = \forall Y \exists Z (y_1 \vee y_2 \vee z_3) \wedge (y_2 \vee \neg z_3 \vee \neg z_4) \wedge (y_2 \vee z_3 \vee z_4)$.

**Reduction 4.** *For a given QBF $\Phi = \forall Y \exists Z \varphi(Y, Z)$, build the AF $F_\Phi^4 = (A_\Phi, R_\Phi)$ with*

$$
\begin{aligned}
A_\Phi &= \{\varphi, \bar\varphi, b, g\} \cup C \cup X \cup \bar X \cup Y' \cup \bar Y' \\
R_\Phi &= \{\langle c, \varphi \rangle \mid c \in C\} \cup \{\langle x, \bar x\rangle, \langle \bar x, x\rangle \mid x \in X\} \cup \\
&\quad \{\langle x, c\rangle, \langle c, x\rangle \mid x \text{ occurs in } c\} \cup \{\langle \bar x, c\rangle, \langle c, \bar x\rangle \mid \neg x \text{ occurs in } c\} \\
&\quad \{\langle y, y'\rangle, \langle y', y\rangle, \langle \bar y, \bar y'\rangle, \langle \bar y', \bar y\rangle \mid y \in Y\} \cup \{\langle \varphi, b\rangle, \langle b, \varphi\rangle, \langle g, g\rangle, \langle g, b\rangle\} \\
&\quad \cup \{(g, y'), (g, \bar y') \mid y \in Y\} \cup \{(\varphi, \bar\varphi), (\bar\varphi, \varphi)\}
\end{aligned}
$$

For an example, see Figure 4.

**Proposition 3.** *For a monotone QBF $\Phi = \forall Y \exists Z \, \varphi$ with each clause containing an literal from $Z \cup \bar Z$, the following statements are equivalent*

1. *$\Phi$ is valid.*

2. *$\varphi$ is skeptically accepted in $F_\Phi^4$ w.r.t. semi-stable semantics.*

3. *$\bar\varphi$ is not credulously accepted in $F_\Phi^4$ w.r.t. semi-stable semantics.*

*Proof sketch.* Notice that this reduction is a variation of the reduction presented in [7] and can be easily shown to be equivalent, using that the arguments $\{g, b\} \cup Y' \cup \bar Y'$ are not acceptable w.r.t. semi-stable semantics. □

**Theorem 4.** *The problem $Cred_{sem}$ is $\Sigma_2^P$-complete and the problem $Skept_{sem}$ is $\Pi_2^P$-complete even for AFs $F$ with $\mathrm{dist}_{SYM}(F) \leq 2$*

*Proof.* By Proposition 3 and the observation that the $F_\Phi^4 - \{\varphi, g\}$ is symmetric. □

**Reduction 5.** *Let $(\varphi, x_\alpha)$ be an instance for the MINSAT problem, i.e. $\varphi$ is a propositional formula over atoms $X$ in CNF and $x_\alpha \in X$. We assume an arbitrary order $<$ on the clauses of $\varphi$. The AF $F_{\varphi,x_\alpha} = (A, R)$ is constructed as follows:*

$$\begin{aligned}
A &= \{\varphi, b, q\} \cup C \cup X \cup \bar{X} \cup \{E_c \mid c \in C\} \\
R &= \{(c, \varphi) \mid c \in C\} \cup \{(\varphi, b), (b, b), (q, x_\alpha)\} \cup \\
&\quad \{(x, \bar{x}), (\bar{x}, x) \mid x \in X\} \cup \\
&\quad \{(l, c) \mid \text{literal } l \text{ occurs in } c \in C\} \cup \\
&\quad \{(E_c, a) \mid c \in C, a \in A \setminus (\{c, \varphi, b\} \cup \{E_{c'} : c' < c\})\}
\end{aligned}$$

**Proposition 4.** *Given $(\varphi, x_\alpha)$ the following statements are equivalent:*

1. *The atom $x_\alpha$ is in a minimal model of $\varphi$.*

2. *The argument $x_\alpha$ is credulously accepted in $F_{\varphi,x_\alpha}$ w.r.t. stage semantics.*

3. *The argument $q$ is not skeptically accepted in $F_{\varphi,x_\alpha}$ w.r.t. stage semantics.*

*Proof.* (1) ⇔ (2) Recall that each stage extension is also a naive extensions, and hence we consider only naive extensions as candidates for stage extensions.

First let us consider naive extensions of $F_{\varphi,x_\alpha} = (A, R)$ containing an argument $E_c$. For simplicity we enumerate the clauses $c_1, \ldots, c_m$ and the arguments $E_1, \ldots, E_m$, according to the order $<$ on the clauses. Now one can easily check that these naive extensions are given by $\{\{E_i, \varphi, q\}, \{E_i, c_i, q\} \mid 1 \leq i \leq m\}$. Further we have that the arguments $E_1, \ldots, E_m$ are in conflict with each other but not attacked from any other argument. Thus when concerning the $\leq_R^+$-maximality of the above naive extensions they only compete with each other but not with any other naive extension. Comparing the range of these extensions we get that stage extensions $E$ such that for some $i$, $E_i \in E$ are the following $\{\{E_i, \varphi, q\} \mid 1 \leq i \leq n\} \cup \{\{E_1, c_1, q\}\}$.

Now let us consider naive sets $E$ such that for each $1 \leq i \leq m$, $E_i \notin E$. As we already have stage extensions with $\{E_i, c_i, q\}^+ = A \setminus \{b\}$ and $\{E_i, \varphi, q\}^+ = A \setminus (\{c_i, E_1, \ldots, E_{i-1}\})$ Clearly $\{E_1, \ldots, E_m\} \cap E = \emptyset$ and thus the only way for $E$ being $\leq_R^+$-maximal is that $\{b, c_1, \ldots, c_m\} \subseteq E^+$. When $b \in E^+$ then we have that $\varphi \in E$ and hence for $1 \leq i \leq m$ $c_i \notin E$. That is that $\{b, c_1, \ldots, c_m\} \subseteq E^+$ iff $\varphi \in E$ and $X \cap E$ is a model of $\varphi$. Hence there is a one-to-one correspondence between models $M$ of $\varphi$ and candidates for stage extensions $M \cup \overline{X \setminus M} \cup \{\varphi\} \cup \{q \mid \text{if } x \notin M\}$ By the construction the range of each this candidate is clearly incomparable with the ranges of the already determined stage extensions $\{\{E_i, \varphi, q\} \mid 1 \leq i \leq n\} \cup \{\{E_1, c_1, q\}\}$. and thus the $\leq_R^+$-maximality of such a candidate only depends on the other candidates.

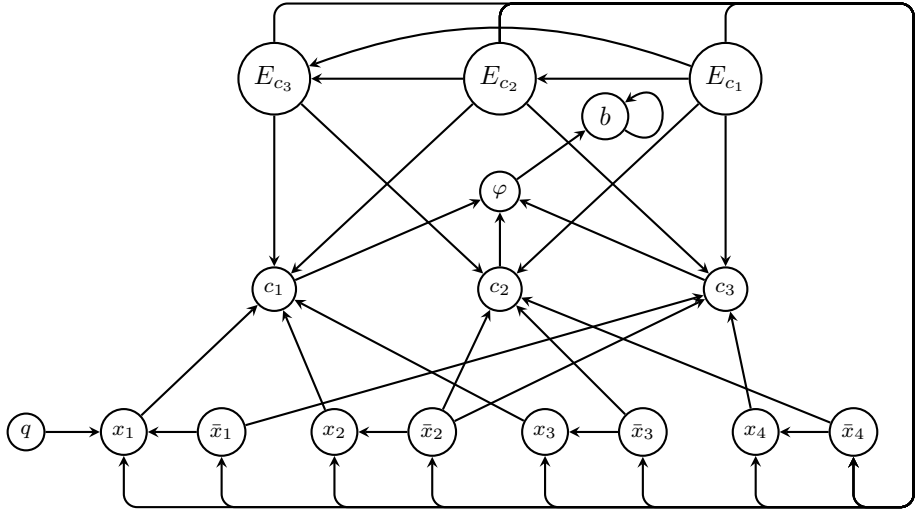

Figure 5: Illustration of the AF $F_{\Phi,x_1}$, for the CNF-formula $\varphi = \bigwedge_{c \in C} c$ with $C = \{\{y_1, y_2, z_3\}, \{\bar{y}_2, \bar{z}_3, \bar{z}_4)\}, \{\bar{y}_1, \bar{y}_2, z_4\}\}$.

It remains to show that for two models $M, N$, $M \subseteq N$ iff $M \cup \overline{X \setminus M} \cup \{\varphi\} \cup \{q \mid \text{if } x \notin M\} \geq_R^+ N \cup \overline{X \setminus N} \cup \{\varphi\} \cup \{q \mid \text{if } x \notin N\}$ For the "only if" direction consider $M \subseteq N$. We have that $(M \cup \overline{X \setminus M} \cup \{\varphi\} \cup \{q \mid \text{if } x \notin M\})^+ = A \setminus (\bar{M} \cup \{E_1, \ldots, E_m\}) \cup \{q \mid \text{if } x \notin M\}$ and $(N \cup \overline{X \setminus N} \cup \{\varphi\} \cup \{q \mid \text{if } x \notin N\})^+ = A \setminus (\bar{N} \cup \{E_1, \ldots, E_m\}) \cup \{q \mid \text{if } x \notin M\}$. As by assumption $M \subseteq N$ we finally have that $A \setminus (\bar{M} \cup \{E_1, \ldots, E_m\}) \supseteq A \setminus (\bar{N} \cup \{E_1, \ldots, E_m\})$. For the "if" part let us consider $M \not\subseteq N$. Hence there is some $x \in M$ such that $x \notin N$. But then we have that $\bar{x} \notin (M \cup \overline{X \setminus M} \cup \{\varphi\} \cup \{q \mid \text{if } x \notin M\})^+$ and $\bar{x} \in (N \cup \overline{X \setminus N} \cup \{\varphi\} \cup \{q \mid \text{if } x \notin N\})^+$. That is that $M \cup \overline{X \setminus M} \cup \{\varphi\} \cup \{q \mid \text{if } x \notin M\} \not\geq_R^+ N \cup \overline{X \setminus N} \cup \{\varphi\} \cup \{q \mid \text{if } x \notin N\}$.

(2) $\Leftrightarrow$ (3): As $x_\alpha$ is the only argument which has a conflict with $q$ we have that each naive extensions, and thus also each stage extension, either contains $q$ or $x_\alpha$. Hence if $q$ is in all stage extensions then $x_\alpha$ is not credulously accepted and vice versa. $\square$

**Theorem 5.** *The problem $Cred_{stg}$ is $\Sigma_2^P$-complete and the problem $Skept_{stg}$ is $\Pi_2^P$-complete even for AFs without even-cycles.*

*Proof.* Immediate by Proposition 4, the fact that Minsat is $\Pi_2^P$-complete [8] and the fact that $F_{\Phi,x_\alpha}$ has no even length cycle. □

**Theorem 6.** *The problem $Cred_{stg}$ is $\Sigma_2^P$-complete and the problem $Skept_{stg}$ is $\Pi_2^P$-complete even for AFs $F$ with $\mathrm{dist}_{ACY}(F) = 1$.*

*Proof.* Immediate by Proposition 4, the fact that Minsat is $\Pi_2^P$-complete [8] and the fact that $F_{\Phi,x_\alpha}$ contains just one cycle, i.e. the self-attacking argument $b$. □

**Reduction 6.** *Given a $QBF_\forall^2$ formula $\Phi = \forall Y \exists Z \varphi$, we define $F_\Phi^5 = (A, R)$, where*

$$\begin{aligned}
A &= \{\varphi, \bar{\varphi}, b, u, v\} \cup C \cup Y \cup \bar{Y} \cup Y' \cup \bar{Y}' \cup Z \cup \bar{Z} \\
R &= \{(c, \varphi) \mid c \in C\} \cup \{(\varphi, \bar{\varphi}), (\bar{\varphi}, \varphi), (\varphi, b), (b, b)\} \cup \\
&\quad \{(x, \bar{x}), (\bar{x}, x) \mid x \in Y \cup Z\} \cup \\
&\quad \{(y, y'), (\bar{y}, \bar{y}'), (y', y), (\bar{y}', \bar{y}) \mid y \in Y\} \cup \\
&\quad \{(l, c), (c, l) \mid \text{literal } l \text{ occurs in } c \in C\} \cup \\
&\quad \{(u, v), (v, v)\} \cup \{(y', u), (\bar{y}', u) \mid y \in Y\}
\end{aligned}$$

**Proposition 5.** *Given a (monotone) $QBF_\forall^2$ formula $\Phi = \forall Y \exists Z \varphi$ the following statements are equivalent:*

1. *$\Phi$ is valid*
2. *$\varphi$ is skeptically accepted in $F_\Phi^5$ w.r.t. stage semantics.*
3. *$\bar{\varphi}$ is not credulously accepted in $F_\Phi^5$ w.r.t. stage semantics.*

*Proof sketch.* Mention that $F_\Phi^5$ is a modification of the reduction presented in [7].

(1) $\Leftrightarrow$ (2) The candidates for being stage extension are $\subseteq$-maximal conflict-free sets, we notate them as $naive(F_\Phi^5)$. We have two classes of naive extensions, those containing the argument $u$ and those not containing $u$, which are incomparable w.r.t. $\leq_R^+$ which can be seen as follows. For $E$ in first class we have $v \in E^+$ and either $y_1' \notin E^+$ or $\bar{y}_1' \notin E^+$ while for $E$ in the latter class $v \notin E^+$ and $y_1', \bar{y}_1' \in E^+$.

So let us first consider the extensions containing $u$ then we have that all of the arguments $Y' \cup \bar{Y}$ are unacceptable and we end up with the same situation as in

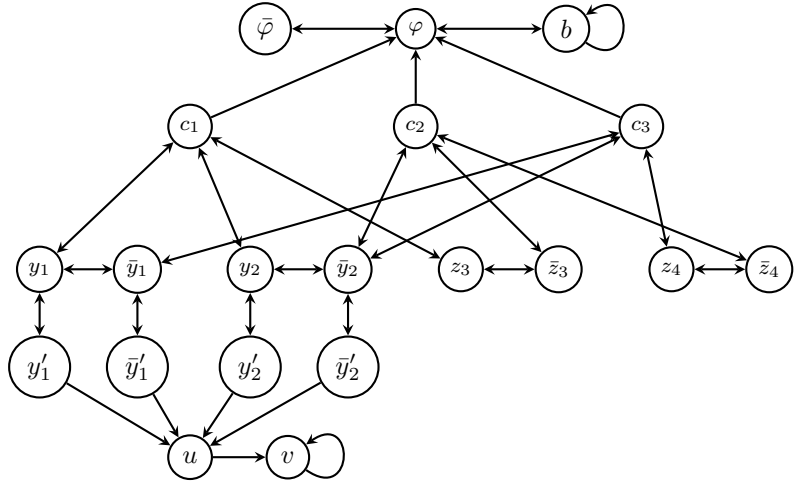

Figure 6: Illustration of the AF $F_\Phi^5$, for $\Phi = \forall y_1 y_2 \exists z_3 z_4 \{\{y_1, y_2, z_3\}, \{\bar{y}_2, \bar{z}_3, \bar{z}_4)\}, \{\bar{y}_1, \bar{y}_2, z_4\}\}$.

[7]. Using the results from [7] we have that $\Phi$ is valid iff $\varphi$ is skeptically accepted in the stage extensions containing $u$.

Now let us consider the extensions $E$ with $u \notin E$. If $\Phi$ is invalid the $\varphi$ fails to be skeptically accepted in the first case anyway. Thus we restrict ourselves to the case where $\Phi$ is valid. Then $\varphi$ has at least one model $M \subseteq Y \cup Z$ and we can construct the extension $S = M \cup \overline{(Y \cup Z) \setminus M} \cup ((Y \cup Z) \setminus M)' \cup \bar{M}' \cup \{\varphi\}$. As $M$ is model of $\varphi$ we have that $S^+ = A \setminus \{v\}$. As no stage extension $E$ with $u \notin E$ satisfies $v \in E^+$, each of these extensions has $E^+ = S^+ = A \setminus \{v\}$. By $b \in E^+$ and $\varphi$ being the only attacker of $b$ we obtain $\varphi \in E$ for each E. Hence $\varphi$ is skeptically accepted.

(2) $\Leftrightarrow$ (3) as $\bar{\varphi}$ is only in conflict with $\varphi$ for each $E \in stg(F_\Phi^5)$ either $\varphi \in E$ or $\bar{\varphi} \in E$ holds. □

**Theorem 7.** *The problem $Cred_{stg}$ is $\Sigma_2^P$-complete and the problem $Skept_{stg}$ is $\Pi_2^P$-complete even for AFs $F$ with* $\text{dist}_{BIP}(F) \leq 4$

*Proof.* By Proposition 5 and the observation that for monotone QBFs the $F_\Phi^5 - \{u, v, b, \varphi\}$ is bipartite. □

**Theorem 8.** *The problem $Cred_{stg}$ is $\Sigma_2^P$-complete and the problem $Skept_{stg}$ is $\Pi_2^P$-complete even for AFs $F$ with* $\text{dist}_{SYM}(F) \leq 4$

*Proof.* By Proposition 5 and the observation that the $F_\Phi^5 - \{u, v, b, \varphi\}$ is symmetric. □